\documentclass[10pt,twocolumn,letterpaper]{article}

\usepackage{cvpr}
\usepackage{times}
\usepackage{epsfig}
\usepackage{graphicx}
\usepackage{amsmath}
\usepackage{amssymb}
\usepackage{mmstyles}
\usepackage{multirow}
\usepackage{pifont}
\usepackage{subcaption}
\usepackage{authblk}

\newcommand{\negative}{\scalebox{0.5}[1.0]{$-$}}
\graphicspath{{figures/}}
\DeclareGraphicsExtensions{.pdf}

\usepackage[breaklinks=true,bookmarks=false]{hyperref}

\cvprfinalcopy % *** Uncomment this line for the final submission

\def\cvprPaperID{2956} % *** Enter the CVPR Paper ID here

\begin{document}

%%%%%%%%% TITLE
%\title{Scene De-occlusion: Ordering Recovery, Amodal and Content Completion \\without Corresponding Annotations}
\title{Self-Supervised Scene De-occlusion}

\author[1]{Xiaohang Zhan}
\author[1]{Xingang Pan}
\author[1]{Bo Dai}
\author[1]{Ziwei Liu}
\author[1]{Dahua Lin}
\author[2]{Chen Change Loy}
\affil[1]{CUHK - SenseTime Joint Lab, The Chinese University of Hong Kong}
\affil[2]{Nanyang Technological University}
\affil[1]{\tt\small \{zx017, px117, bdai, zwliu, dhlin\}@ie.cuhk.edu.hk}
\affil[2]{\tt\small ccloy@ntu.edu.sg}

\twocolumn[{
	\renewcommand\twocolumn[1][]{#1}
	\maketitle
	\begin{center}
		\centering
		\includegraphics[width=0.9\linewidth]{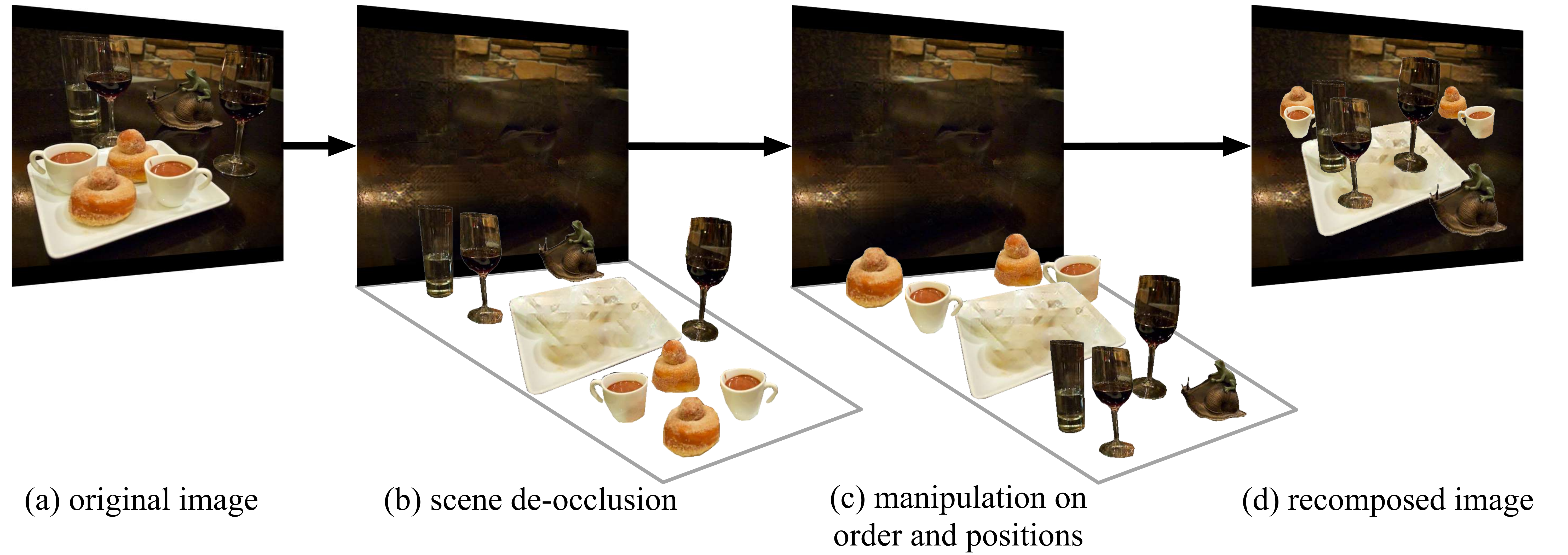}
		\captionof{figure}{Scene de-occlusion decomposes an image, extracting cluttered objects in it into entities of indiviual intact objects. Orders and positions of the extracted objects can be manipulated to recompose new scenes.}
		\label{fig:recomp}
	\end{center}
}]

\maketitle
%\thispagestyle{empty}

% !TEX root = ../cameraready.tex

\begin{abstract}

Natural scene understanding is a challenging task, particularly when encountering images of multiple objects that are partially occluded. This obstacle is given rise by varying object ordering and positioning. Existing scene understanding paradigms are able to parse only the visible parts, resulting in incomplete and unstructured scene interpretation. In this paper, we investigate the problem of scene de-occlusion, which aims to recover the underlying occlusion ordering and complete the invisible parts of occluded objects. We make the first attempt to address the problem through a novel and unified framework that recovers hidden scene structures without ordering and amodal annotations as supervisions. This is achieved via Partial Completion Network (PCNet)-mask (M) and -content (C), that learn to recover fractions of object masks and contents, respectively, in a self-supervised manner. Based on PCNet-M and PCNet-C, we devise a novel inference scheme to accomplish scene de-occlusion, via progressive ordering recovery, amodal completion and content completion. Extensive experiments on real-world scenes demonstrate the superior performance of our approach to other alternatives. Remarkably, our approach that is trained in a self-supervised manner achieves comparable results to fully-supervised methods. The proposed scene de-occlusion framework benefits many applications, including high-quality and controllable image manipulation and scene recomposition (see Fig.~\ref{fig:recomp}), as well as the conversion of existing modal mask annotations to amodal mask annotations. Project page: \url{https://xiaohangzhan.github.io/projects/deocclusion/}.

\end{abstract}

\vspace{-0.5cm}
% !TEX root = ../cameraready.tex

\section{Introduction}
\label{sec:intro}

Scene understanding is one of the foundations of machine perception. 
A real-world scene, regardless of its context, often comprises multiple objects of varying ordering and positioning, with one or more object(s) being occluded by other object(s).
Hence, scene understanding systems should be able to process modal perception, \ie, parsing the directly visible regions, as well as amodal perception~\cite{kanizsa1979organization,palmer1999vision,lehar1999gestalt}, \ie, perceiving the intact structures of entities including invisible parts.
The advent of advanced deep networks along with large-scale annotated datasets has facilitated many scene understanding tasks, \eg, object detection~\cite{felzenszwalb2010cascade,ren2015faster,Wang_2019_CVPR,Chen_2018_CVPR}, scene parsing~\cite{liu2015semantic,chen2017deeplab,zhao2017pyramid}, and instance segmentation~\cite{dai2016instance,he2017mask,chen2019hybrid,Wang_2019_ICCV}.
Nonetheless, these tasks mainly concentrate on modal perception, while amodal perception remains rarely explored to date.

A key problem in amodal perception is \textit{scene de-occlusion}, which involves the subtasks of recovering the underlying occlusion ordering and completing the invisible parts of occluded objects.
While human vision system is capable of intuitively performing scene de-occlusion, elucidation of occlusions is highly challenging for machines.
First, the relationships between an object that occludes other object(s), called an ``occluder'', and an object that is being occluded by other object(s), called an ``occludee'', is profoundly complicated.
This is especially true when there are multiple ``occluders'' and ``occludees'' with high intricacies between them, namely an ``occluder'' that occludes multiple ``occludees'' and an ``ocludee'' that is occluded by multiple ``occluders'', forming a complex occlusion graph.
Second, depending on the category, orientation, and position of objects, the boundaries of ``occludee(s)'' are elusive; no simple priors can be applied to recover the invisible boundaries.

A possible solution for scene de-occlusion is to train a model with \emph{ground truth} of occlusion orderings and amodal masks (\ie, intact instance masks).
Such ground truth can be obtained either from synthetic data \cite{ehsani2018segan,hu2019sail} or from manual annotations on real-world data \cite{zhu2017semantic,qi2019amodal,follmann2019learning}, each of which with specific limitations.
The former introduces inevitable domain gap between the fabricated data used for training and the real-world scene in testing.
The latter relies on subjective interpretation of individual annotators to demarcate occluded boundaries, therefore subjected to biases, and requires repeated annotations from different annotators to reduce noise, therefore are laborious and costly.
A more practical and scalable way is to learn scene de-occlusion from the data itself rather than annotations.
%
%Previous attempts~\cite{wu2017neural,burgess2019monet} have succeeded in this line of research, but is restricted to performing scene de-occlusion on images of artificially rendered 2D or 3D objects.

\begin{figure}[t]
	\centering  
	\includegraphics[width=\linewidth]{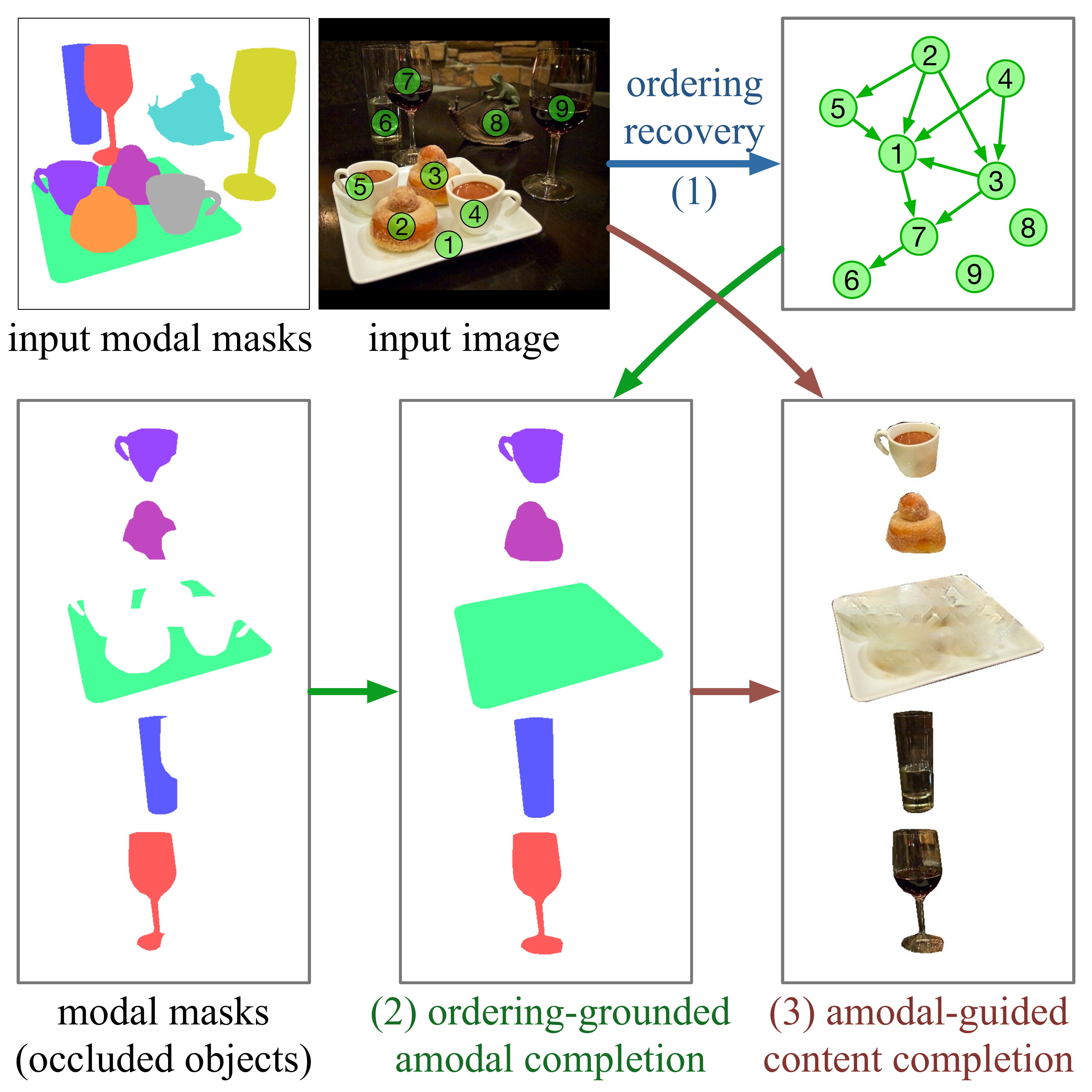}
	\caption{Given an input image and the associated modal masks, our framework solves scene de-occlusion progressively -- 1) predicts occlusion ordering between different objects as a directed graph, 2) performs amodal completion grounded on the ordering graph, and 3) furnishes the occluded regions with content under the guidance of amodal predictions. The de-occlusion is achieved by two novel networks, PCNet-M and PCNet-C, which are trained without annotations of ordering or amodal masks.}
	\label{fig:intro}
\end{figure}

In this work, we propose a novel self-supervised framework that tackles scene de-occlusion on real-world data \textit{without manual annotations of occlusion ordering or amodal masks}.
In the absence of ground truth, an end-to-end supervised learning framework is not applicable anymore.
We therefore introduce a unique concept of \emph{partial completion} of occluded objects.
There are two core precepts in the partial completion notion that enables attainment of scene de-occlusion in a self-supervised manner.
First, the process of completing an ``occludee'' occluded by multiple ``occluders'' can be broken down into a sequence of partial completions, with one ``occluder'' involved at a time.
Second, the learning of making partial completion can be achieved by further trimming down the ``occludee'' deliberately and training a network to recover the previous untrimmed occludee.
We show that partial completion is sufficient to complete an occluded object progressively, as well as to facilitate the reasoning of occlusion ordering.

Partial completion is executed via two networks, \ie, Partial Completion Network-mask and -content. We abbreviate them as PCNet-M and PCNet-C, respectively.
PCNet-M is trained to \emph{partially} recover the invisible mask of the ``occludee'' corresponding to an occluder, while PCNet-C is trained to \emph{partially} fill in the recovered mask with RGB content.
%%
%%PCNets serve as primary\xh{fundamental?} components in our de-occlusion framework.
%%
%
%The training is conducted without any ordering or amodal annotations.
%
PCNet-M and PCNet-C form the two core components of our framework to address scene de-occlusion.
%The process is illustrated in Fig.~\ref{fig:intro}.

As illustrated in Fig.~\ref{fig:intro}, the proposed framework takes a real-world scene and its corresponding modal masks of objects, derived from either annotations or predictions of existing modal segmentation techniques, as inputs.
Our framework then streamlines three subtasks to be tackled progressively:
\textbf{1) Ordering Recovery.} Given a pair of neighboring objects in which one can be occluding the other, following the principle that PCNet-M partially completes the mask of the ``occludee'' while keeping the ``occluder'' unmodified, the roles of the two objects are determined.
We recover the ordering of all neighboring pairs and obtain a directed graph that captures the occlusion order among all objects.
\textbf{2) Amodal Completion.} For a specific ``occludee'', the ordering graph indicates all its ``occluders''. Grounded on this information and reusing PCNet-M, an amodal completion method is devised to fully complete the modal mask into an amodal mask of the ``occludee''.
\textbf{3) Content Completion.} The predicted amodal mask indicates the occluded region of an ``occludee''. Using PCNet-C, we furnish RGB content into the invisible region.
With such a progressive framework,
we decompose a complicated scene into isolated and intact objects, along with a highly accurate occlusion ordering graph, allowing subsequent manipulation on the ordering and positioning of objects to recompose a new scene, as shown in Fig.~\ref{fig:recomp}.

We summarize our contributions as follows:
\textbf{1)} We streamline scene de-occlusion into three subtasks, namely ordering recovery, amodal completion, and content completion.
\textbf{2)} We propose PCNets and a novel inference scheme to perform scene de-occlusion without the need for corresponding manual annotations. Yet, we observe comparable results to fully-supervised approaches on datasets of real scenes.
\textbf{3)} The self-supervised nature of our approach shows its potential to endow large-scale instance segmentation datasets, \eg, KITTI~\cite{geiger2013vision}, COCO~\cite{lin2014microsoft}, \etc, with high-accuracy ordering and amodal annotations.
\textbf{4)} Our scene de-occlusion framework represents a novel enabling technology for real-world scene manipulation and recomposition, providing a new dimension for image editing.
% !TEX root = ../cameraready.tex

\section{Related Work}
\begin{figure*}[t]
	\centering
	\includegraphics[width=\linewidth]{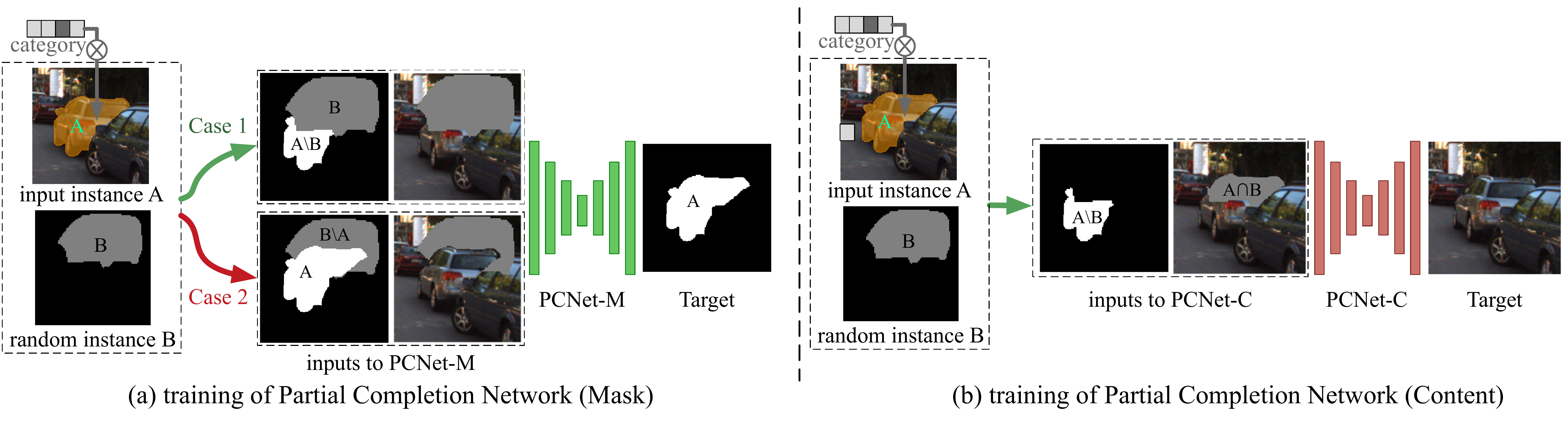}
	\vskip -0.2cm
	\caption{The training procedure of the PCNet-M and the PCNet-C. Given an instance A as the input, we randomly sample another instance B from the whole dataset and position it randomly. Note that we only have modal masks of both A and B. (a) PCNet-M is trained by switching two cases. Case 1 (A erased by B) follows the partial completion mechanism where PCNet-M is encouraged to partially complete A. Case 2 prevents PCNet-M from over completing A. (b) PCNet-C uses $A\cap B$ to erase A and learn to fill in the RGB content of the erased region. It also takes in $A\backslash B$ as an additional input. The modal mask of $A$ is multiplied with its category id if available.}
	\label{fig:train_framework}
	%\vspace{-0.1cm}
\end{figure*}

\noindent\textbf{Ordering Recovery.}
In the unsupervised stream, Wu~\etal~\cite{wu2017neural} propose to recover ordering by re-composing the scene with object templates.
However, they only demonstrate the system on toy data.
Tighe~\etal~\cite{tighe2014scene} build a prior occlusion matrix between classes on the training set and minimize quadratic programming to recover the ordering in testing. The inter-class occlusion prior ignores the complexity of realistic scenes.
Other works~\cite{hoiem2007recovering,purkait2019seeing} rely on additional depth cues.
However, depth is not reliable in occlusion reasoning, \eg, there is no depth difference if a piece of paper lies on a table.
The assumption made by these works that farther objects are occluded by close ones also does not always hold. For example, as shown in Fig.~\ref{fig:intro}. The plate (\#1) is occluded by the coffee cup (\#5), while the cup is farther in depth.
In the supervised stream, several works manually annotate occlusion ordering~\cite{zhu2017semantic,qi2019amodal} or rely on synthetic data~\cite{hu2019sail} to learn the ordering in a fully-supervised manner.
Another stream of works on panoptic segmentation~\cite{liu2019end,lazarow2019learning} design end-to-end training procedures to resolve overlapping segments. However, they do not explicitly recover the full scene ordering.

\noindent\textbf{Amodal Instance Segmentation.}
Modal segmentation, such as semantic segmentation~\cite{chen2017deeplab,zhao2017pyramid} and instance segmentation~\cite{dai2016instance,he2017mask,chen2019hybrid}, aims at assigning categorical or object labels to visible pixels.
Existing approaches for modal segmentation are not able to solve the de-occlusion problem.
Different from modal segmentation, amodal instance segmentation aims at detecting objects as well as recovering the amodal (integrated) masks of them.
Li~\etal~\cite{li2016amodal} produces dummy supervision through pasting artificial occluders, while the absence of explicit ordering increases the difficulty when complicated occlusion relationship is present.
Other works take a fully-supervised learning approach by using either manual annotations~\cite{zhu2017semantic,qi2019amodal,follmann2019learning} or synthetic data~\cite{hu2019sail}.
As mentioned above, it is costly and inaccurate to annotate invisible masks manually.
Approaches relying on synthetic data are also confronted with domain gap issues.
On the contrary, our approach can convert modal masks into amodal masks in a self-supervised manner. This unique ability facilitates the training of amodal instance segmentation networks without manual amodal annotations.

\noindent\textbf{Amodal Completion.}
Amodal completion is slightly different from amodal instance segmentation.
In amodal completion, modal masks are given at test time and the task is to complete the modal masks into amodal masks.
Previous works on amodal completion typically rely on heuristic assumptions on the invisible boundaries to perform amodal completion with given ordering relationships.
Kimia~\etal~\cite{kimia2003euler} propose to adopt Euler Spiral in amodal completion.
Lin~\etal~\cite{lin2016computational} use cubic B\'{e}zier curves.
Silberman~\etal~\cite{silberman2014contour} apply curve primitives including straight lines and parabolas.
Since these studies still require ordering as the input, they cannot be adopted directly to solve de-occlusion problem.
Besides, these unsupervised approaches mainly focus on toy examples with simple shapes.
Kar~\etal~\cite{kar2015amodal} use keypoint annotations to align 3D object templates to 2D image objects, so as to generate the ground truth of amodal bounding boxes.
Ehsani~\etal~\cite{ehsani2018segan} leverage 3D synthetic data to train an end-to-end amodal completion network.
Similar to unsupervised methods, our framework does not need annotations of amodal masks or any kind of 3D/synthetic data.
In contrast, our approach is able to solve amodal completion in highly cluttered natural scenes, whereas other unsupervised methods fall short.

\if 0
\noindent\textbf{Unsupervised Scene decomposition.}
%
Previous works~\cite{wu2017neural,burgess2019monet} propose to decompose scenes into isolated objects in estimated ordering with reconstruction as the supervision.
%
These works focus on toy data and never scale up.
%
By contrast, our approach is able to perform scene decomposition on realistic scenes without corresponding annotations.
\fi

% !TEX root = ../cameraready.tex

\section{Our Scene De-occlusion Approach}
% overview
The proposed framework aims at \textbf{1)} recovering occlusion ordering and \textbf{2)} completing amodal masks and content of occluded objects.
To cope with the absence of manual annotations of occlusion ordering and amodal masks, we design a way to train the proposed PCNet-M and PCNet-C to complete instances partially in a self-supervised manner.
With the trained networks, we further propose a progressive inference scheme to perform ordering recovery, ordering-grounded amodal completion, and amodal-constrained content completion to complete objects.

\subsection{Partial Completion Networks (PCNets)}
Given an image, it is easy to obtain the modal masks of objects via off-the-shelf instance segmentation frameworks.
However, their amodal masks are unavailable.
Even worse, we do not know whether these modal masks are intact, making the learning of full completion of an occluded instance extremely challenging.
The problem motivates us to explore self-supervised partial completion.

\noindent
\textbf{Motivation.}
Suppose an instance's modal mask constitutes a pixel set $M$, we denote the ground truth amodal mask as $G$.
Supervised approaches solve the full completion problem of $M\xrightarrow[]{f_\theta} G$, where $f_\theta$ denotes the full completion model.
This full completion process can be broken down into a sequence of partial completions $M\xrightarrow[]{p_\theta} M_1\xrightarrow[]{p_\theta} M_2\xrightarrow[]{p_\theta} \cdots\xrightarrow[]{p_\theta} G$ if the instance is occluded by multiple ``occluders'', where $M_k$ is the intermediate states, $p_\theta$ denotes the partial completion model.

Since we still do not have any ground truth to train the partial completion step $p_\theta$, we take a step back by further trimming down $M$ randomly to obtain $M_{\negative 1}$ s.t. $M_{\negative 1}\subset M$.
Then we train $p_\theta$ via $M_{\negative 1}\xrightarrow[]{p_\theta} M$. %, representing a self-supervised training manner.
The self-supervised partial completion approximates the supervised one, laying the foundation of our PCNets.
Based on such a self-supervised notion, we introduce Partial Completion Networks (PCNets).
They contain two networks, respectively, for mask (PCNet-M) and content completion (PCNet-C).

\noindent
\textbf{PCNet-M for Mask Completion.}
The training of PCNet-M is shown in Fig.~\ref{fig:train_framework} (a). We first prepare the training data. Given an instance A along with its modal mask $M_A$ from the dataset $D$ with instance-level annotations, we randomly sample another instance B from $D$ and position it randomly to acquire a mask $M_B$.
Here we regard $M_A$ and $M_B$ as sets of pixels.
There are two input cases, in which different input is fed to the network: 

\noindent
1) The \textit{first case} corresponds to the aforementioned partial completion strategy.
We define $M_B$ as an eraser,
and use B to erase part of A to obtain $M_{A\backslash B}$.
In this case, the PCNet-M is trained to recover the original modal mask $M_A$ from $M_{A\backslash B}$, conditioned on $M_B$.

\noindent
2) The \textit{second case} serves as a regularization to discourage the network from over-completing an instance if the instance is not occluded.
Specifically, $M_{B\backslash A}$ that does not invade A is regarded as the eraser.
In this case, we encourage the PCNet-M to retain the original modal mask $M_A$, conditioned on $M_{B\backslash A}$.
Without case 2, the PCNet-M always encourage increment of pixels, which may result in over-completion of an instance if it is not occluded by other neighboring instances.

In both cases, the erased image patch serves as an auxiliary input.
We formulate the loss functions as follows:
\begin{equation}
\small{
\begin{split}
&L_1 = \frac{1}{N}\sum_{A,B \in D} L\left(P_\theta^{\text{(m)}} \left(M_{A\backslash B}~;  M_B,I\backslash M_B\right), M_A\right), \\
&L_2 = \frac{1}{N}\sum_{A,B \in D} L\left(P_\theta^{\text{(m)}} \left(M_A~; M_{B\backslash A}, I\backslash M_{B\backslash A} \right), M_A\right),
\end{split}
}
\end{equation}
where $P_\theta^{\text{(m)}} \left(\star\right)$ is our PCNet-M network, $\theta$ represents the parameters to optimize, $I$ is the image patch, $L$ is Binary Cross-Entropy Loss.
We formulate the final loss function as $L^\text{(m)} = x L_1 + (1-x) L_2, x\sim Bernoulli\left(\gamma\right)$, 
where $\gamma$ is the probability to choose case 1.
The random switching between the two cases forces the network to understand the ordering relationship between the two neighboring instances from their shapes and border, so as to determine whether to complete the instance or not.

\noindent
\textbf{PCNet-C for Content Completion.}
PCNet-C follows a similar intuition of PCNet-M, while the target to complete is RGB content.
As shown in Fig.~\ref{fig:train_framework} (b), the input instances A and B are the same as that for PCNet-M.
Image pixels in region $M_{A\cap B}$ are erased, and PCNet-C aims at predicting the missing content.
Besides, PCNet-C also takes in the remaining mask of A, \ie, $M_{A\backslash B}$ to indicate that it is A rather than other objects, that is painted.
Hence, it cannot be simply replaced by standard image inpainting approaches.
The loss of PCNet-C to minimize is formulated as follows:
\begin{equation}
\small{
	L^\text{(c)} = \frac{1}{N}\sum_{A,B \in D} L\left(P_\theta^{\text{(c)}} \left(I\backslash M_{A\cap B}; M_{A\backslash B}, M_{A\cap B}\right), I\right),
	}
\end{equation}
where $P_\theta^{\text{(c)}}$ is our PCNet-C network, $I$ is the image patch, $L$ represents the loss function consisting of common losses in image inpainting including $l_1$, perceptual and adversarial loss.
Similar to PCNet-M, the training of PCNet-C via learning partial completion enables full completion of the instance content at test time.

\begin{figure}[t]
	\centering
	\includegraphics[width=\linewidth]{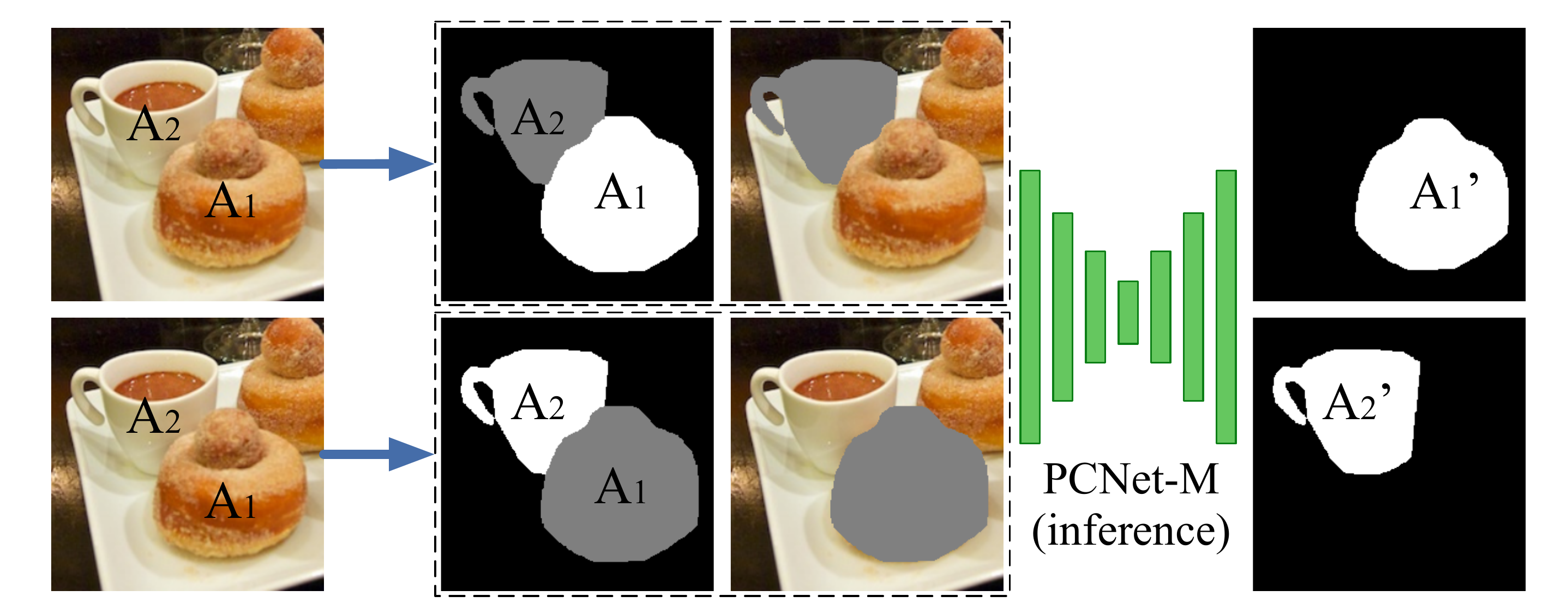}
	\caption{Dual-Completion for ordering recovery. To recover the ordering between a pair of neighboring instances $A_1$ and $A_2$, we switch the role of the target object (in white) and the eraser (in gray). The increment of $A_2$ is larger than that of $A_1$, thus $A_2$ is identified as the ``occludee''.}
	\label{fig:dualcomp}
\end{figure}

\subsection{Dual-Completion for Ordering Recovery}
The target ordering graph is composed of pair-wise occlusion relationships between all neighboring instance pairs.
A neighboring instance pair is defined as two instances whose modal masks are connected, thus one of them possibly occludes the other.
As shown in Fig.~\ref{fig:dualcomp}, given a pair of neighboring instances $A_1$ and $A_2$, we first regard $A_1$'s modal mask $M_{A_1}$ as the target to complete. $M_{A_2}$ serves as the eraser to obtain the increment of $A_1$, \ie, $\Delta_{A_1|A_2}$.
Symmetrically, we also obtain the increment of $A_2$ conditioned on $A_1$, \ie, $\Delta_{A_2|A_1}$.
The instance gaining a larger increment in partial completion is supposed to be the ``occludee''.
Hence, we infer the order between $A_1$ and $A_2$ via comparing their incremental area, as follows:
\begin{equation}
\small{
\begin{split}
	& \Delta_{A_1|A_2} = P_\theta^{\text{(m)}}\left(M_{A_1}~;~ M_{A_2},I\backslash M_{A_2}\right)~\backslash~M_{A_1},\\
	& \Delta_{A_2|A_1} = P_\theta^{\text{(m)}}\left(M_{A_2}~;~ M_{A_1},I\backslash M_{A_1}\right)~\backslash~M_{A_2}, \\
	& O\left(A_1, A_2\right) =
	\begin{cases}
		0, & \text{if~} |\Delta_{A_1|A_2}| = |\Delta_{A_2|A_1}| = 0 \\
		1, & \text{if~} |\Delta_{A_1|A_2}| < |\Delta_{A_2|A_1}| \\
		\negative 1, & \text{otherwise}
	\end{cases},
\end{split}
}	
\end{equation}
where $O\left(A_1, A_2\right)=1$ indicates that $A_1$ occludes $A_2$.
If $A_1$ and $A_2$ are not neighboring, $O\left(A_1, A_2\right) = 0$.
Note that in practice the probability of $\left|\Delta_{A_1|A_2}\right| = \left|\Delta_{A_2|A_1}\right| > 0$ is zero, thus does not need to be specifically considered here.
Performing Dual-Completion for all neighboring pairs provides us the scene occlusion ordering, which can be represented as a directed graph as shown in Fig.~\ref{fig:intro}.
The nodes in the graph represent objects, while edges indicate the directions of occlusion between neighboring objects.
Note that it is not necessarily to be acyclic, as shown in Fig.~\ref{fig:interlock}.

\begin{figure}[t]
	\centering
	\includegraphics[width=.87\linewidth]{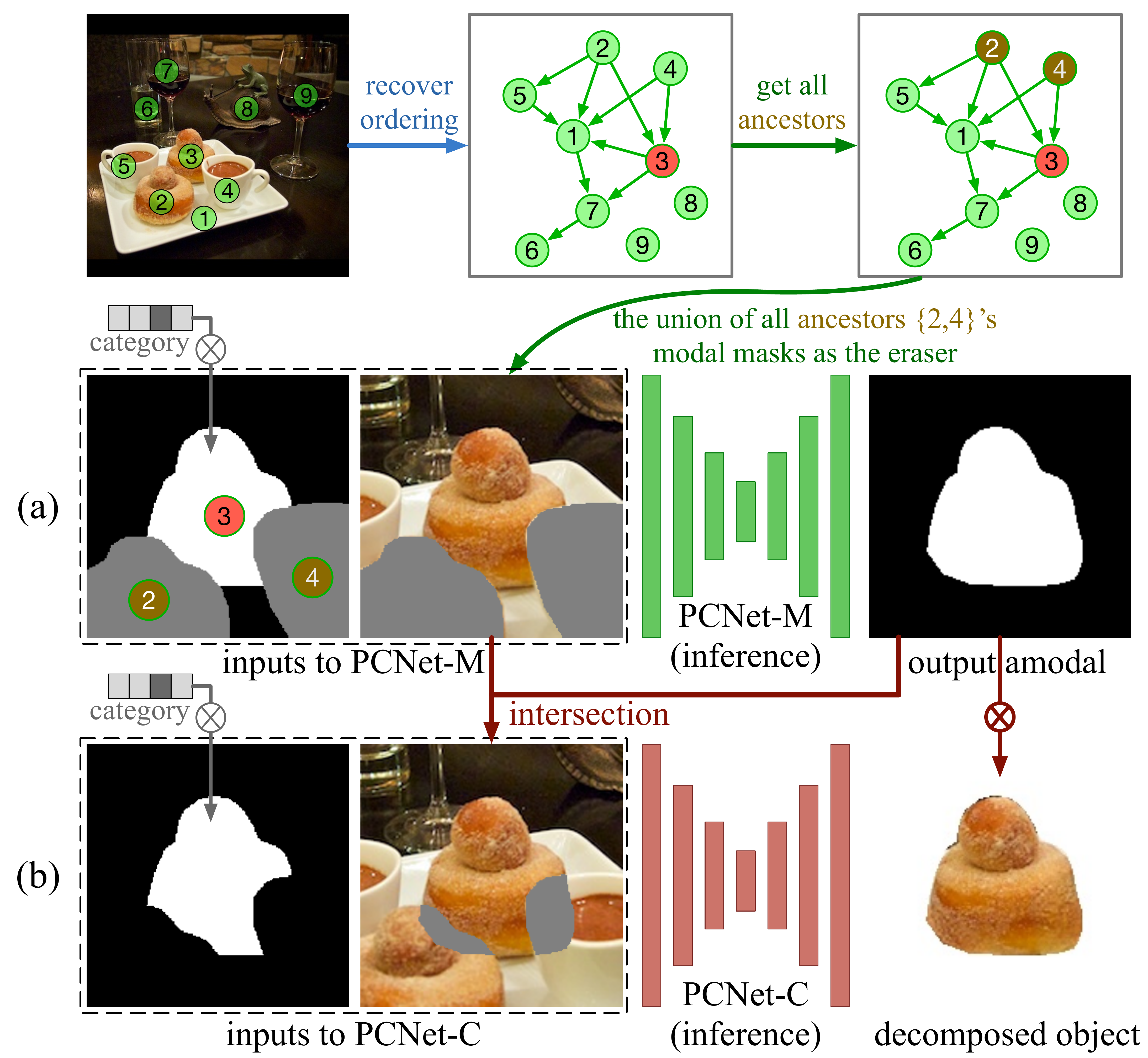}
	\caption{(a) Ordering-grounded amodal completion takes the modal mask of the target object (\#3) and all its ancestors (\#2, \#4), as well as the erased image as inputs. With the trained PCNet-M, it predicts the amodal mask of object \#3. (b) The intersection of the amodal mask and the ancestors indicates the invisible region of object \#3. Amodal-constrained content completion (red arrows) adopts the PCNet-C to fill in the content in the invisible region.}
	\label{fig:completion}
\end{figure}

\subsection{Amodal and Content Completion}

\begin{figure}[t]
	\centering
	\includegraphics[width=\linewidth]{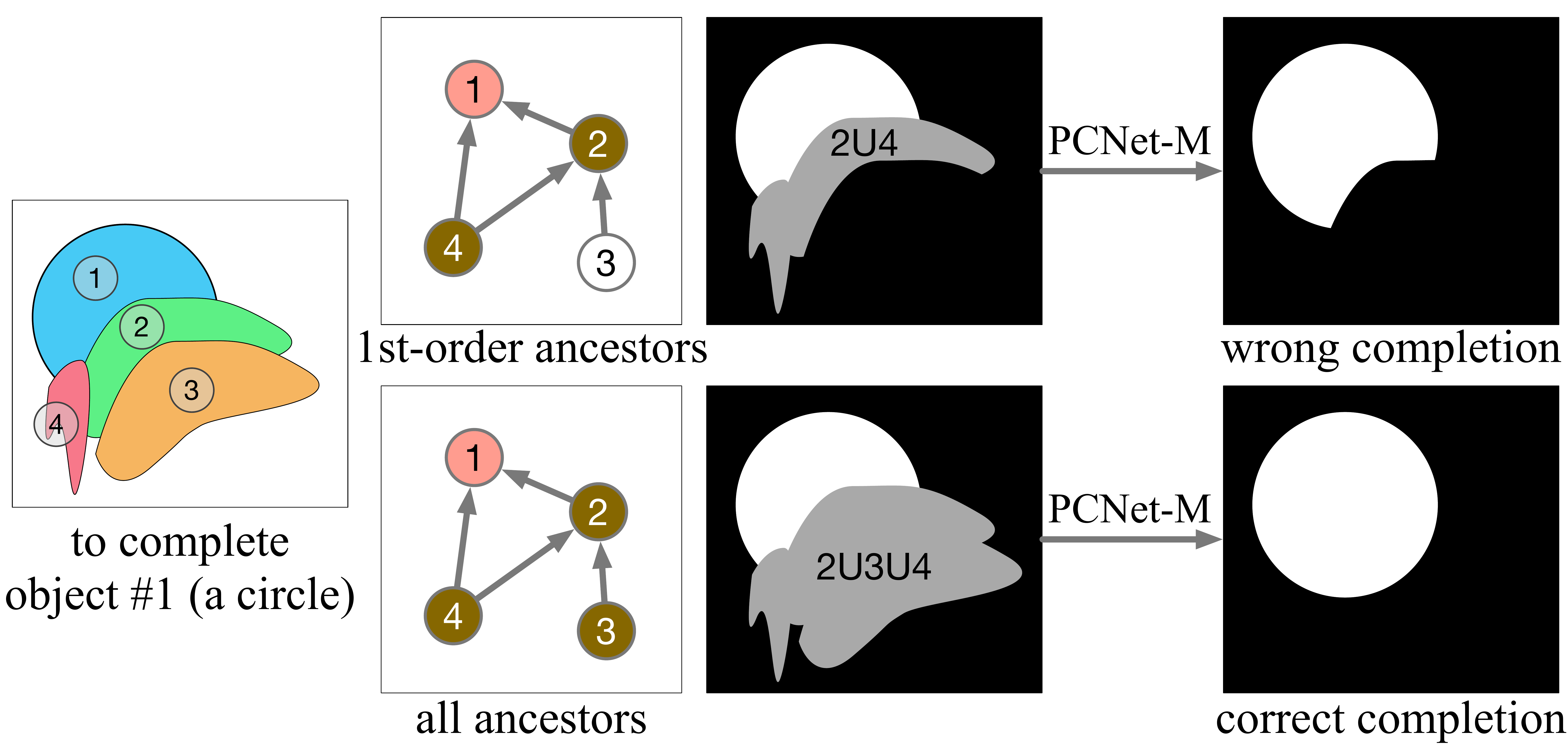}
	\caption{This figure shows why we need to find all ancestors rather than only the first-order ancestors, though higher-order ancestors do not directly occlude this instance. Higher-order ancestors (\eg, instance \#3) may indirectly occlude the target instance (\#1), thus need to be taken into account.}
	\label{fig:ancestor}
%	\vspace{-0.2cm}
\end{figure}

\noindent\textbf{Ordering-Grounded Amodal Completion.}
We can perform ordering-grounded amodal completion after estimating the ordering graph.
%With the estimated ordering graph, we propose Ordering-Grounded Amodal Completion algorithm.
%
Suppose we need to complete an instance $A$, we first find all ancestors of $A$ in the graph as the ``occluders'' of this instance via breadth-first searching (BFS).
Since the graph is not necessarily to be acyclic, we adapt the BFS algorithm accordingly.
Interestingly, we find that the trained PCNet-M is generalizable to use the union of all ancestors as the eraser.
Hence, we do not need to iterate the ancestors and apply PCNet-M to partially complete $A$ step by step.
Instead, we perform amodal completion in one step conditioned on the union of all ancestors' modal masks.
Denoting the ancestors of $A$ as $\{\text{anc}_i^A, i=1,2,\cdots,k\}$, we perform amodal completion as follows:
\begin{equation}
\small{
\begin{split}
	&Am_A = P_\theta^{\text{(m)}}\left(M_A~;~M_{\text{anc}^A},I\backslash M_{\text{anc}^A}\right), \\
	&M_{\text{anc}^A} = \bigcup_{i=1}^k M_{\text{anc}_i^A},
\end{split}
}
\end{equation}
where $Am_A$ is the result of amodal mask, $M_{\text{anc}_i^A}$ is the modal mask of $i$-th ancestor.
An example is shown in Fig.~\ref{fig:completion} (a).
Fig.~\ref{fig:ancestor} shows the reason we use all ancestors rather than only the first-order ancestor.

\noindent\textbf{Amodal-Constrained Content Completion.}
In previous steps, we obtain the occlusion ordering graph and the predicted amodal mask of each instance.
Next, we complete the occluded content of them.
As shown in Fig.~\ref{fig:completion} (b), the intersection of predicted amodal mask and the ancestors $Am_A\cap M_{\text{anc}^A}$ indicates the missing part of $A$, regarded as the eraser for PCNet-C.
Then we apply a trained PCNet-C to fill in the content as follows:
\begin{equation}
\small{
\begin{split}
	&C_A = P_\theta^{\text{(c)}}\left(I\backslash M_E~;~M_A,M_E\right)\circ Am_A, \\
	&M_E = Am_A\cap M_{\text{anc}^A},
\end{split}
}
\end{equation}
where $C_A$ is the decomposed content of $A$ from the scene.
For background contents, we use the union of all foreground instances as the eraser.
Different from image inpainting that is unaware of occlusion, content completion is performed on the estimated occluded regions.

% !TEX root = ../cameraready.tex

\section{Experiments}
We now evaluate our method in various applications including \textit{ordering recovery}, \textit{amodal completion}, \textit{amodal instance segmentation}, and \textit{scene manipulation}.
The implementation details and more qualitative results can be found in the supplementary materials.

\noindent\textbf{Datasets.}
\textbf{1) KINS}~\cite{qi2019amodal}, originated from KITTI~\cite{geiger2013vision}, is a large-scale traffic dataset with annotated modal and amodal masks of instances. PCNets are trained on the training split (7,474 images, 95,311 instances) with modal annotations. We test our de-occlusion framework on the testing split (7,517 images, 92,492 instances).
\textbf{2) COCOA}~\cite{zhu2017semantic} is a subset of COCO2014~\cite{lin2014microsoft} while annotated with pair-wise ordering, modal, and amodal masks. We train PCNets on the training split (2,500 images, 22,163 instances) using modal annotations and test on the validation split (1,323 images, 12,753 instances).
The categories of instance are unavailable for this dataset. Hence, we set the category id constantly as 1 in training PCNets for this dataset.
%\textbf{3) LVIS}~\cite{gupta2019lvis} is a large-scale long-tail fine-grained instance dataset. There are no ordering or amodal annotations for this dataset. We train our PCNets on the training split and show qualitative results on the validation split.

\subsection{Comparison Results}
\begin{figure}[t]
	\centering
	\includegraphics[width=\linewidth]{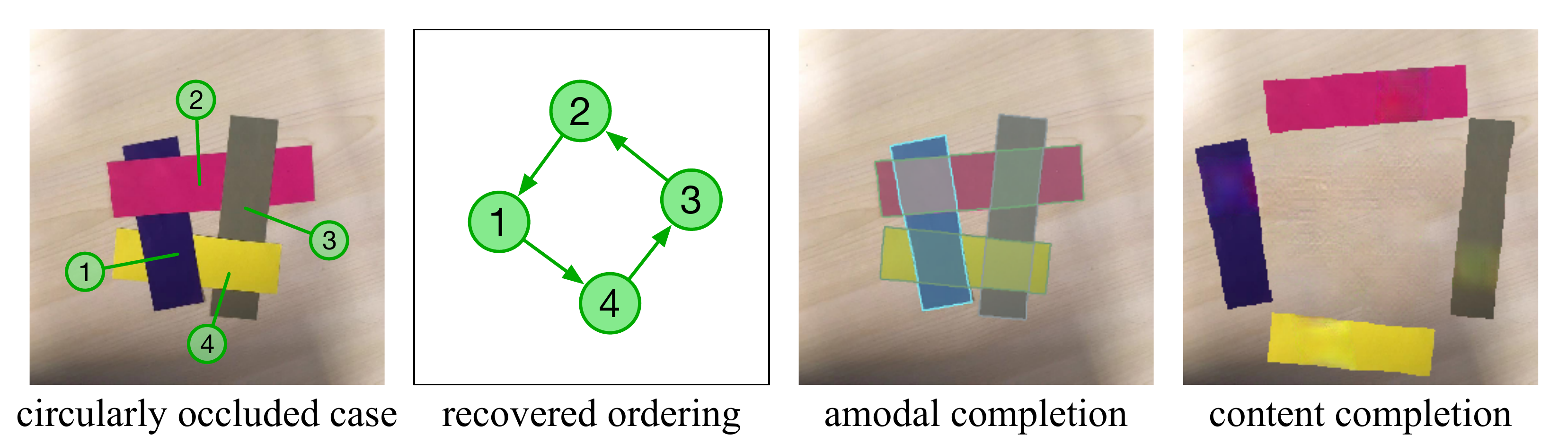}
	\caption{Our framework is able to solve circularly occluded cases. Since such case is rare, we cut four pieces of paper to compose it.}
	\label{fig:interlock}
\end{figure}

\noindent\textbf{Ordering Recovery.} We report ordering recovery performance on COCOA and KINS in Table~\ref{tab:order}.
We reproduced the OrderNet proposed in~\cite{zhu2017semantic} to obtain the supervised results.
Baselines include sorting bordered instance pairs by \emph{Area}\footnote{We optimize this heuristic depending on each dataset -- a larger instance is treated as a front object for KINS, and opposite for COCOA.}, \emph{Y-axis} (instance closer to image bottom in front), and \emph{Convex} prior.
For baseline \emph{Convex}, we compute convex hull on modal masks to approximate amodal completion, and the object with more increments is regarded as the occludee.
All baselines have been adjusted to achieve their respective best performances.
On both benchmarks, our method achieves much higher accuracies than baselines, comparable to the supervised counterparts.
An interesting case is shown in Fig.~\ref{fig:interlock}, where four objects are circularly overlapped.
Since our ordering recovery algorithm recovers pair-wise ordering rather than sequential ordering, it is able to solve this case and recover the cyclic directed graph.

\begin{table}[t]
	\centering
	\caption{Ordering estimation on COCOA validation and KINS testing sets, reported with pair-wise accuracy on occluded instance pairs.}
	\begin{tabular}{l|c|c|c}
		\hline
		method & gt order (train) & COCOA & KINS \\\hline
		\multicolumn{4}{l}{\emph{Supervised}}\\\hline
		OrderNet\textsuperscript{M}~\cite{zhu2017semantic} & \ding{52} & 81.7 & 87.5 \\\hline
		OrderNet\textsuperscript{M+I}~\cite{zhu2017semantic} & \ding{52} & 88.3 & 94.1 \\\hline\hline
		\multicolumn{4}{l}{\emph{Unsupervised}}\\\hline
		Area      & \ding{56} & 62.4   & 77.4 \\\hline
		Y-axis & \ding{56} & 58.7   &  81.9 \\\hline
		Convex     & \ding{56} & 76.0   & 76.3  \\\hline
		Ours     & \ding{56} & \textbf{87.1}   & \textbf{92.5}  \\\hline
	\end{tabular}
	\label{tab:order}
	%\vspace{-0.2cm}
\end{table}

%\vspace{5pt}
\noindent\textbf{Amodal Completion.}
We first introduce the baselines.
For the \emph{supervised} method, amodal annotation is available.
A UNet is trained to predict amodal masks from modal masks end-to-end.
\emph{Raw} means no completion is performed.
\emph{Convex} represents computing the convex hull of the modal mask as the amodal mask.
Since the convex hull usually leads to over-completion, \ie, extending the visible mask, we improve this baseline by using predicted order to refine the convex hull, constituting a stronger baseline: \emph{Convex\textsuperscript{R}}.
It performs pretty well for naturally convex objects.
\emph{Ours (NOG)} represents the non-ordering-grounded amodal completion that relies on our PCNet-M and regards all neighboring objects as the eraser rather than using occlusion ordering to search the ancestors.
\emph{Ours (OG)} is our ordering-grounded amodal completion method.

We evaluate amodal completion on ground truth modal masks, as shown in Table~\ref{tab:amodalcomp_gtmodal}.
Our method surpasses the baseline approaches and are comparable to the supervised counterpart.
The comparison between \emph{OG} and \emph{NOG} shows the importance of ordering in amodal completion.
As shown in Fig.~\ref{fig:amodal_display}, some of our results are potentially more natural than manual annotations.

\begin{table}[t]
	\centering
	\caption{Amodal completion on COCOA validation and KINS testing sets, using ground truth modal masks.}
	\begin{tabular}{l|c|c|c}
		\hline
		method & \begin{tabular}{c}amodal\\(train)\end{tabular}  & \begin{tabular}{c}COCOA\\\%mIoU\end{tabular} & \begin{tabular}{c}KINS\\\%mIoU\end{tabular}\\\hline
		Supervised  & \ding{52} & 82.53 & 94.81 \\
		Raw  	& \ding{56} & 65.47 & 87.03 \\
		Convex\textsuperscript{R}  	& \ding{56} &74.43  & 90.75 \\\hline
		Ours (NOG)  & \ding{56}  & 76.91 & 93.42\\
		Ours (OG)   	& \ding{56} & \textbf{81.35} & \textbf{94.76}  \\\hline\hline
	\end{tabular}
	\label{tab:amodalcomp_gtmodal}
\end{table}

Apart from using ground truth modal masks as the input in testing, we also verify the effectiveness of our approach with predicted modal masks as the input.
Specifically, we train a UNet to predict modal masks from an image.
In order to correctly match the modal and the corresponding ground truth amodal masks in evaluation, we use the bounding box as an additional input to this network.
We predict the modal masks on the testing set, yielding 52.7\% mAP to the ground truth modal masks.
We use the predicted modal masks as the input to perform amodal completion.
As shown in Table~\ref{tab:amodalcomp_predmodal}, our approach still achieves high performance, comparable to the supervised counterpart.

\begin{table}[t]
	\centering
	\caption{Amodal completion on KINS testing set, using predicted modal masks (mAP 52.7\%).}
	\begin{tabular}{l|c|c}
		\hline
		method & amodal (train)  & KINS \%mIoU\\\hline
		Supervised  & \ding{52}   & 87.29 \\
		Raw   	& \ding{56}& 82.05 \\
		Convex\textsuperscript{R} 	& \ding{56}	& 84.12 \\\hline
		Ours (NOG) 	& \ding{56}	& 85.39 \\
		Ours (OG) 	& \ding{56}  & \textbf{86.26} \\\hline
	\end{tabular}
	\label{tab:amodalcomp_predmodal}
\end{table}

\begin{table}[t]
	\centering
	\caption{Amodal instance segmentation on KINS testing set. Convex\textsuperscript{R} means using predicted order to refine the convex hull. In this experimental setting, all methods detect and segment instances from raw images. Hence, modal masks are not used in testing.}
	\begin{tabular}{l|c|c|c}
		\hline
		Ann. source & modal (train) & amodal (train)  & \%mAP \\\hline
		GT~\cite{qi2019amodal}  & \ding{56}& \ding{52}  & 29.3 \\\hline\hline
		%GT (M+A)~\cite{qi2019amodal}  & \ding{52}& \ding{52}   & 31.1 \\\hline
		Raw     & \ding{52} & \ding{56}    & 22.7  \\\hline
		Convex & \ding{52}    & \ding{56}  & 22.2   \\\hline
		Convex\textsuperscript{R}  & \ding{52}    & \ding{56}  & 25.9   \\\hline
		Ours & \ding{52} & \ding{56}  & \textbf{29.3}  \\\hline
	\end{tabular}
	\label{tab:amodalseg}
\end{table}

\begin{figure}[t]
	\centering
	\includegraphics[width=\linewidth]{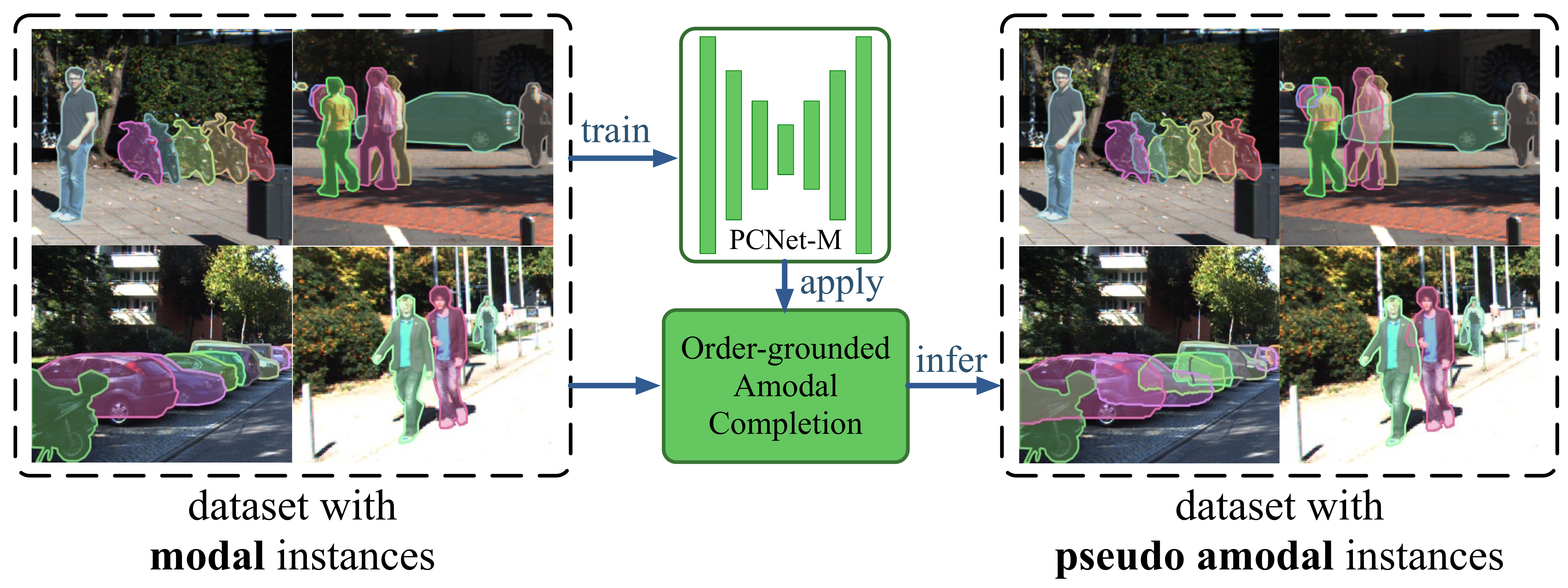}
	\caption{By training the self-supervised PCNet-M on a modal dataset (\eg, KITTI shown here) and applying our amodal completion algorithm on the same dataset, we are able to freely convert modal annotations into pseudo amodal annotations. Note that such self-supervised conversion is intrinsically different from training a supervised model on a small labeled amodal dataset and applying it to a larger modal dataset, where the generalizability between different datasets can be an issue.}
	\label{fig:trainingset_convert}
	%\vspace{-0.1cm}
\end{figure}

\begin{figure*}[t]
	\centering
	\includegraphics[width=\linewidth]{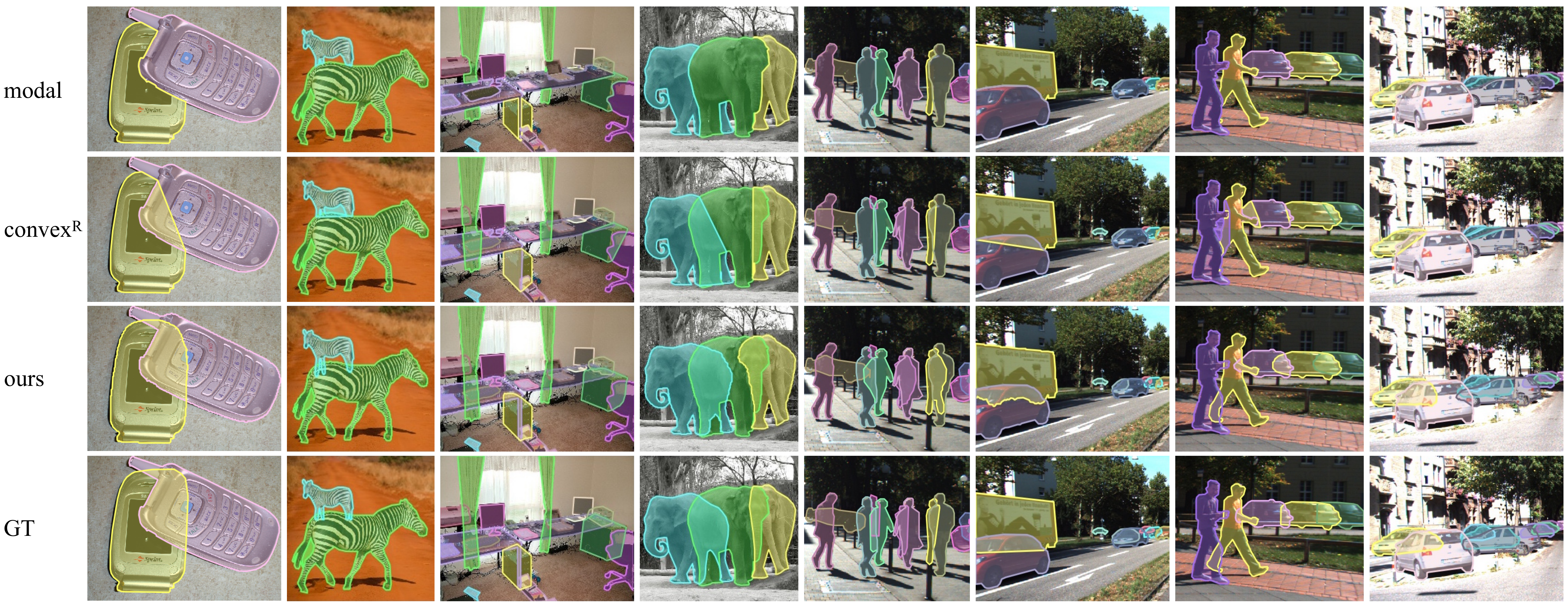}
	\caption{Amodal completion results. Our results are potentially more natural than manual annotations (GT) in some cases, especially for instances in yellow.}
	\label{fig:amodal_display}
\end{figure*}

\begin{figure}[t]
	\centering
	\includegraphics[width=\linewidth]{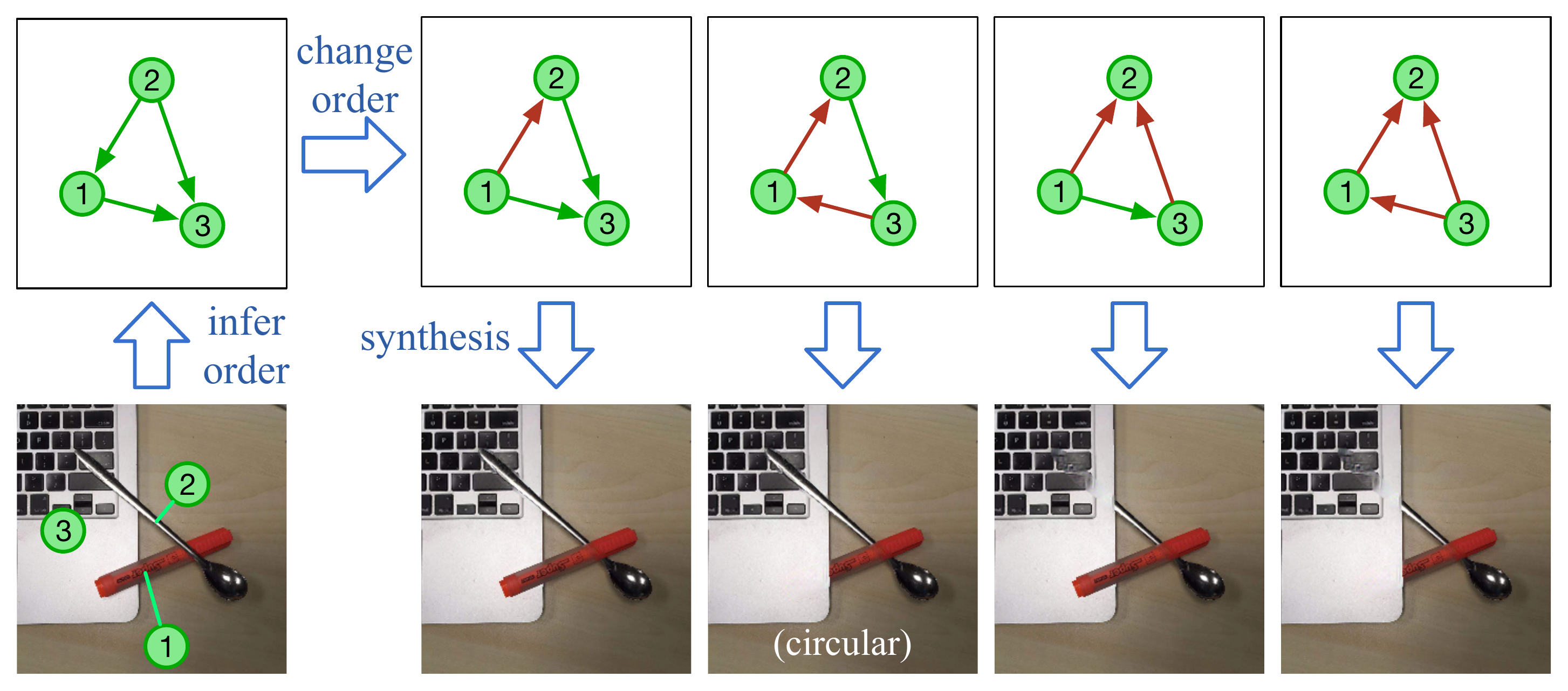}
	\caption{Scene synthesis by changing the ordering graph. Reversed orderings are shown in red arrows. Uncommon cases with circular ordering can also be synthesized.}
	\label{fig:orderchange}
	%\vspace{-0.1cm}
\end{figure}

\begin{figure}[t]
	\centering
	\includegraphics[width=\linewidth]{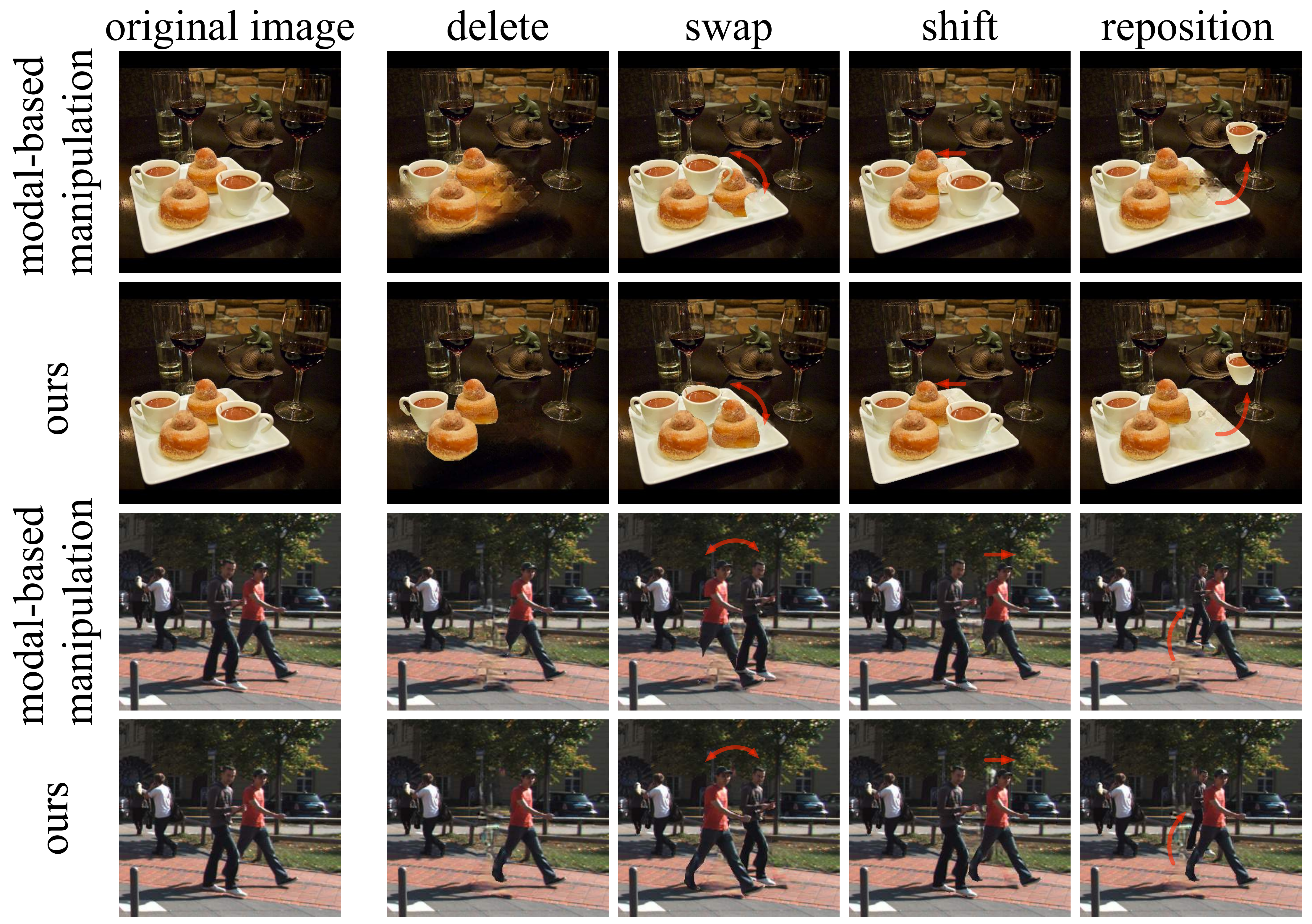}
	\caption{This figure shows rich and high-quality manipulations, including deleting, swapping, shifting and repositioning instances, enabled by our approach. The baseline method \emph{modal-based manipulation} is based on image inpainting, where modal masks are provided, order and amodal masks are unknown. Better in zoomed-in view. More examples can be found in the supplementary material.}
	\label{fig:manpulate}
	\vspace{-4pt}
\end{figure}

%\vspace{5pt}
\noindent\textbf{Label Conversation for Amodal instance segmentation.}
Amodal instance segmentation aims at detecting instances and predicting amodal masks from images simultaneously.
With our approach, one can convert an existing dataset with modal annotations into the one with \textit{pseudo amodal annotations}, thus allowing amodal instance segmentation network training without manual amodal annotations.
This is achieved by training PCNet-M on the modal mask training split, and applying our amodal completion algorithm on the same training split to obtain the corresponding amodal masks, as shown in Fig.~\ref{fig:trainingset_convert},
%
%In this way, our framework is able to freely convert a dataset with modal annotations into the one with pseudo-amodal ``annotations''.
%
To evaluate the quality of the pseudo amodal annotations, we train a standard Mask R-CNN~\cite{he2017mask} for amodal instance segmentation following the setting in~\cite{qi2019amodal}.
All baselines follow the same training protocol, except that the amodal annotations for training are different.
As shown in Table~\ref{tab:amodalseg}, using our inferred amodal bounding boxes and masks, we achieve the same performance (mAP 29.3\%) as the one using manual amodal annotations.
Besides, our inferred amodal masks in the training set are highly consistent with the manual annotations (mIoU 95.22\%).
The results suggest a high applicability of our method for obtaining reliable pseudo amodal mask annotations, relieving burdens of manual annotation on large-scale instance-level datasets.

\subsection{Application on Scene Manipulation}
Our scene de-occlusion framework allows us to decompose a scene into the background and isolated completed objects, along with an occlusion ordering graph.
Therefore, manipulating scenes by controlling order and positions is made possible.
Fig.~\ref{fig:orderchange} shows scene synthesis by controlling order only.
Fig.~\ref{fig:manpulate} shows more manipulation cases, indicating that our de-occlusion framework, though trained without any extra information compared to the baseline, enables high-quality \emph{occlusion-aware manipulation}.

% !TEX root = ../cameraready.tex

\section{Conclusion}

To summarize, we have proposed a unified scene de-occlusion framework equipped with self-supervised PCNets trained without ordering or amodal annotations.
The framework is applied in a progressive way to recover occlusion orderings, then perform amodal and content completion.
It achieves comparable performances to the fully-supervised counterparts on real-world datasets.
It is applicable to convert existing modal annotations to amodal annotations.
Quantitative results show their equivalent efficacy to manual annotations.
Furthermore, our framework enables high-quality occlusion-aware scene manipulation, providing a new dimension for image editing.

\small{\noindent\textbf{Acknowledgement}: This work is supported by the SenseTime-NTU Collaboration Project, Collaborative Research grant from SenseTime Group (CUHK Agreement No. TS1610626 \& No. TS1712093), Singapore MOE AcRF Tier 1 (2018-T1-002-056), NTU SUG, and NTU NAP.}

\clearpage

{\small
\bibliographystyle{ieee_fullname}
\bibliography{bib}
}

\appendix
\section{Implementation Details}
In our experiments, the backbone for PCNet-M is UNet~\cite{ronneberger2015u} with a widening factor 2, and that for PCNet-C is a UNet equipped with partial convolution layers~\cite{liu2018partialinpainting}; while note that PCNets do not have restrictions on backbone architectures.
For both PCNets, the image or mask patches centering on an object are cropped by an adaptive square and resized to 256x256 as inputs.

For COCOA, the PCNet-M is trained using SGD for 56K iterations with an initial learning rate 0.001 decayed at iterations 32K and 48K by 0.1.
For KINS, we stop the training process earlier at 32K.
The batch size is 256 distributed on 8 GPUs (GTX 1080 TI).
The hyper-parameter $\gamma$ that balances the two cases in training PCNet-M is set to $0.8$.
In current experiments, we do not use RGB as an input to PCNet-M, since we empirically find that introducing RGB through concatenation makes little differences.
It is probably because for these two datasets, modal masks are informative enough for training; while we believe in more complicated scenes, RGB will exert more influence if introduced in a better way.

For PCNet-C, we modify the UNet to take in the concatenation of image and modal mask as the input.
Apart from the losses in ~\cite{liu2018partialinpainting}, we add an extra adversarial loss for optimization.
The discriminator is a stack of 5 convolution layers with spectral normalization and leaky ReLU (slope=$0.2$).
The PCNet-C is fine-tuned for 450K iterations with a constant learning rate $10^{-4}$ from a pre-trained inpainting network~\cite{liu2018partialinpainting}.
We adapt the pre-trained weights to be compatible for taking in the additional modal mask.

\section{Discussions}

\subsection{Analysis on varying occlusion ratio.}

\begin{figure}[t]
	\centering
	\includegraphics[width=0.85\linewidth]{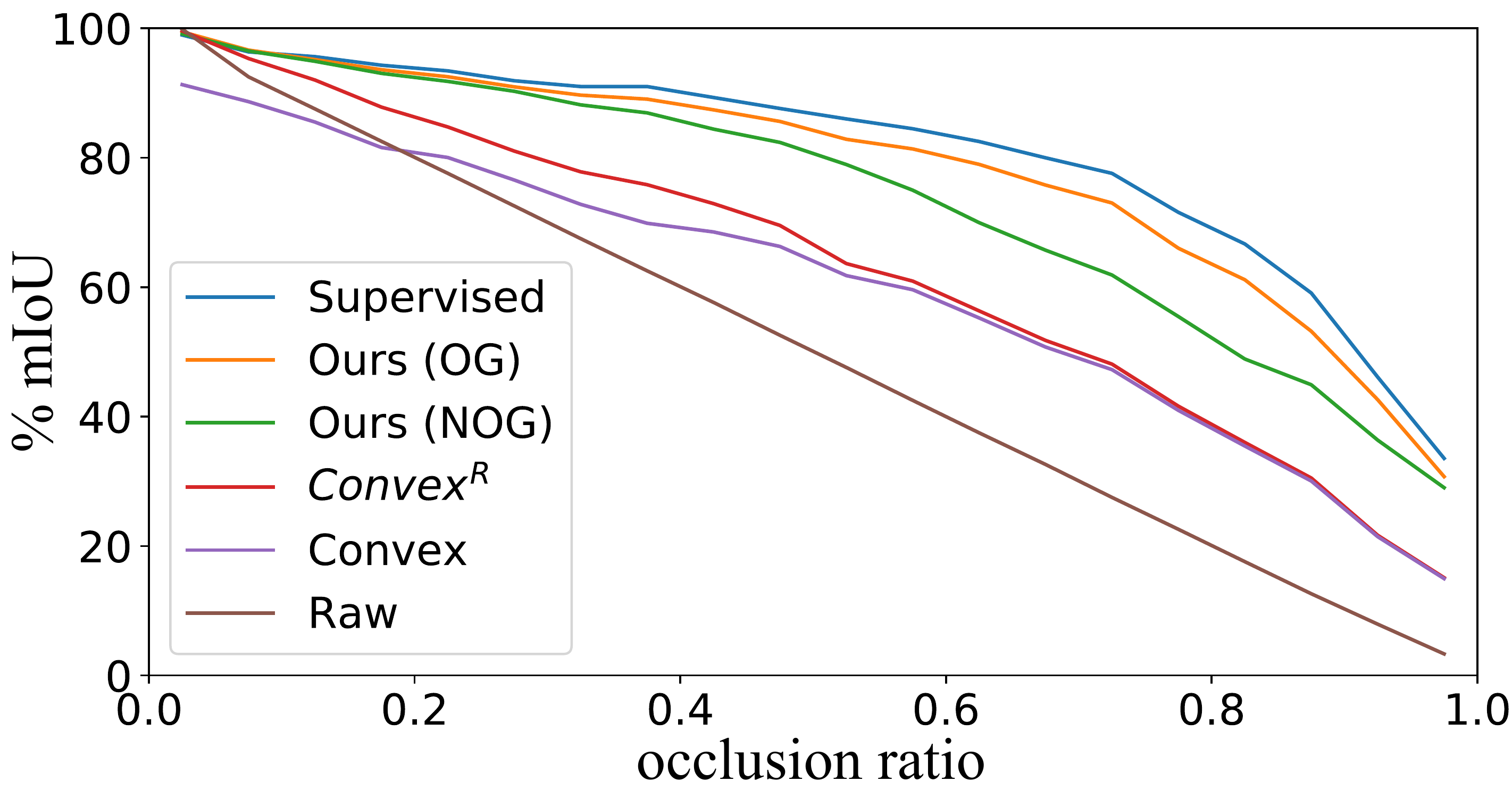}
	\caption{Performances of different approaches under a growing occlusion ratio, evaluated on KINS testing set.}
	\label{fig:analysis}
\end{figure}

Fig.~\ref{fig:analysis} show the amodal completion performances of different approaches under varying ratios of occluded area.
Naturally, larger occlusion ratios result in lower performances.
Under high occlusion ratios, our full method (\emph{Ours (OG)}) surpasses the baseline methods by a large margin.

\subsection{Does it support mutual occlusion?}

\begin{figure}[t]
	\centering
	\includegraphics[width=\linewidth]{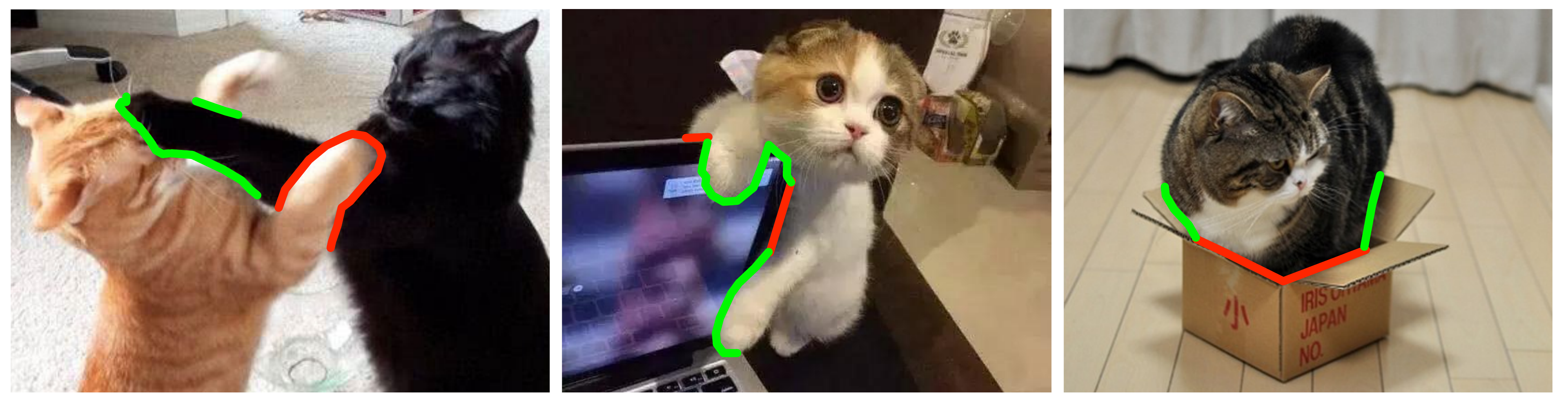}
	\caption{Mutual occlusion cases. Green boundaries show one object occlude the other and red boundaries vice versa.}
	\label{fig:mutual}
\end{figure}

As a drawback, our approach does not support cases where two objects are mutually occluded as shown in ~\ref{fig:mutual}, because our approach focuses on object-level de-occlusion.
For mutual occlusions, the ordering graph cannot be defined, therefore fine-grained boundary-level de-occlusion is required.
It leaves an open question to scene de-occlusion problem.
Nonetheless, our approach works well if more than two objects are cyclically occluded as shown in Fig. 7 in the main paper.

\subsection{Will case 2 mislead PCNet-M?}

\begin{figure}[t]
	\centering
	\includegraphics[width=\linewidth]{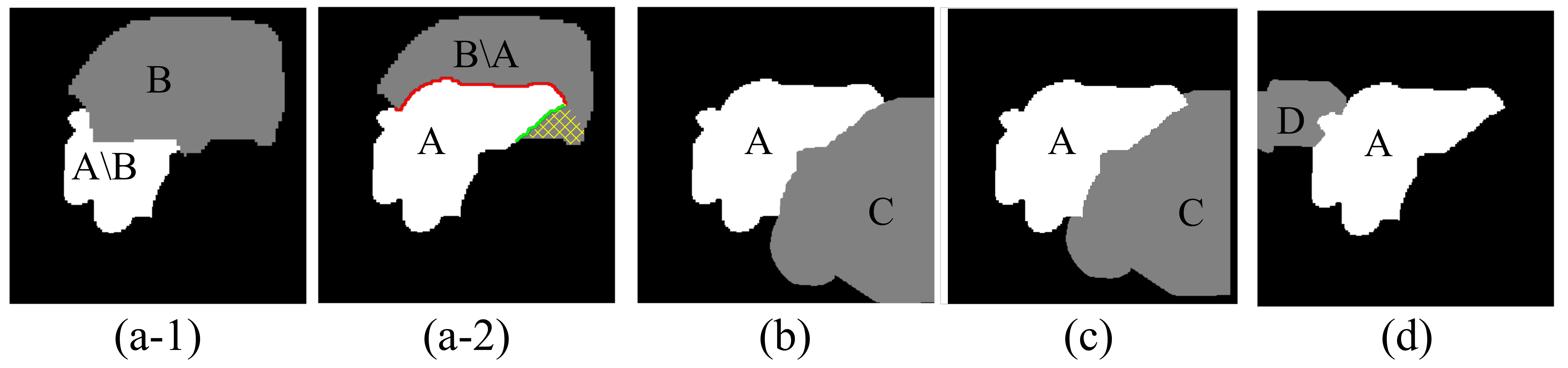}
	\caption{(a-1) and (a-2) represent case 1 and case 2 in training, respectively; (b) - (d) represent possible cases in testing. Among the test cases, only the A in (b) will be completed.}
	\label{fig:discussion}
\end{figure}

As shown in Fig.\ref{fig:discussion}, one may have concerns that in case (a-2) when not-to-complete strategy is applied, the boundary between $A$ and $B\backslash A$ might include a contour shown in green where $A$ is occluded by a real object, namely $C$.
Therefore, it might teach PCNet-M a wrong lesson if the yellow shaded region is taught not to be filled.

Here we explain why it will not teach PCNet-M the wrong lesson.
First of all, PCNet-M learns to complete or not to complete the target object \emph{conditioned on} a surrogate occluder.
As shown in Fig.~\ref{fig:discussion}, as PCNet-M is taught to complete $A\backslash B$ in (a-1) while not to complete $A$ in (a-2), it has to discover cues indicating that $A$ is below $B$ in (a-1) and $A$ is above $B$ in (a-2).
The cues might include the shape of two objects, the shape of common boundary, junctions, \etc.
In testing time, \eg in (b) when regarding the real $C$ as the condition, it is easy for PCNet-M to tell that $C$ is above $A$ from those cues. Therefore PCNet-M actually inclines to case 1, when $A$ will be completed conditioned on $C$.

Then which case does this not-to-complete strategy affect?
The case in (c) shares very similar occlusion patterns with (a-2), especially in the upper right part of the common boundary, showing strong cues that $A$ is above $C$, in which case PCNet-M will not complete $A$ as expected.
However, case (c) is abnormal and unlikely to exist in the real world.
The situation where the not-to-complete strategy really takes effect lies in case (d).
In this case when strong cues indicate that $A$ is above $D$, the PCNet-M is taught not to extend $A$ across $A\&D$ boundary to invade $D$.

\section{Visualization}
\begin{figure*}[t]
	\centering
	\includegraphics[width=\linewidth]{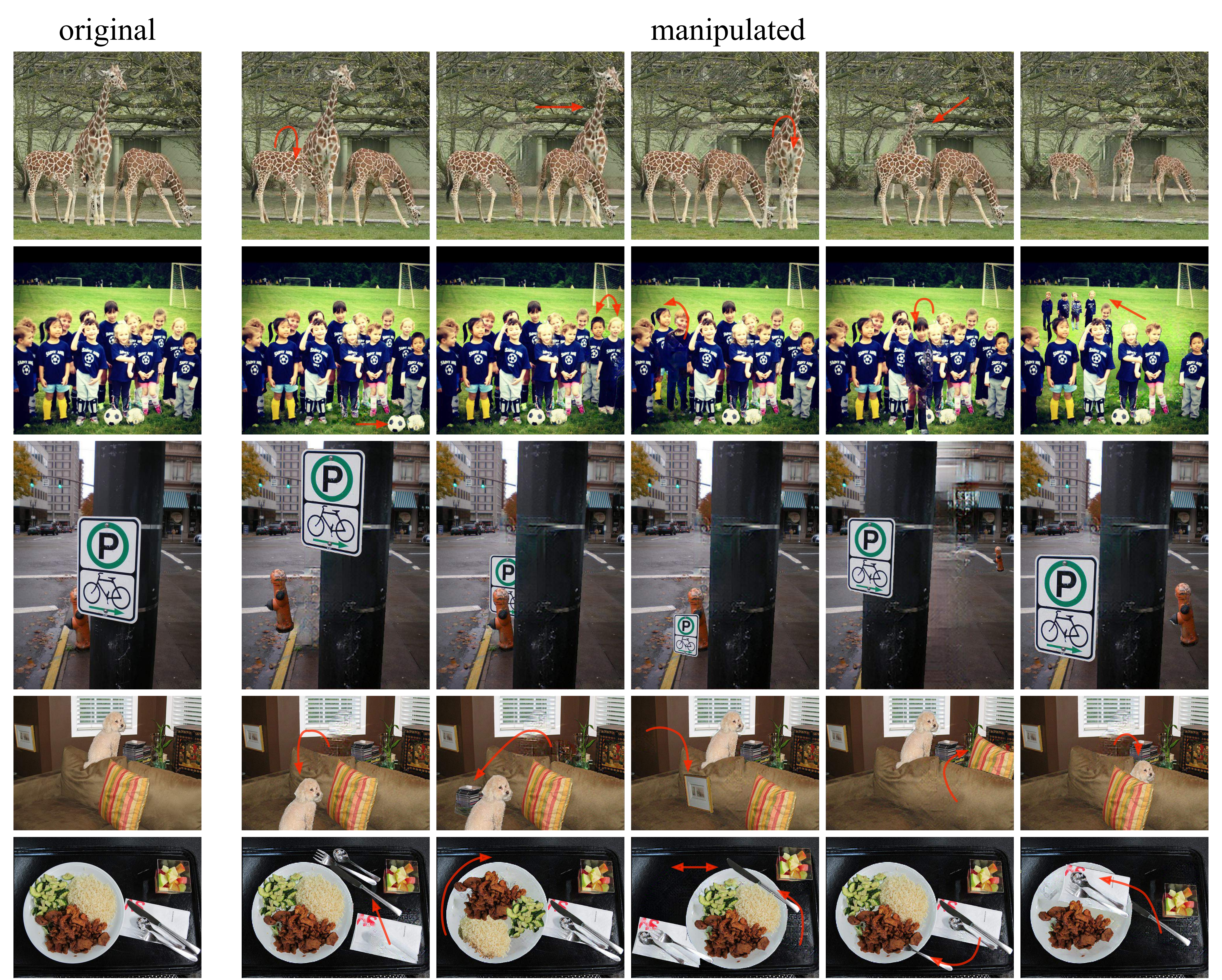}
	\caption{Scene manipulation results based on our de-occlusion framework. Inconspicuous changes are marked with red arrows. A video demo can be found in the project page: \url{https://xiaohangzhan.github.io/projects/deocclusion/}.}
	\label{fig:manipulate_supp}
\end{figure*}

As shown in Fig.~\ref{fig:manipulate_supp}, our approach enables us to freely adjust scene spatial configurations to re-compose new scenes.
The quality could be further improved with the advance of image inpainting, since the PCNet-C shares a similar network architecture and training strategy to image inpainting.

\end{document}